# Using Large Language Models for the Interpretation of Building Regulations

Stefan Fuchs*, The University of Auckland, New Zealand
Michael Witbrock, The University of Auckland, New Zealand
Johannes Dimyadi, CAS Limited, The University of Auckland, New Zealand
Robert Amor, The University of Auckland, New Zealand

*sffc348@aucklanduni.ac.nz (Corresponding Author)

___________________________________________________________________________

**Abstract:** Compliance checking is an essential part of a construction project. The recent rapid uptake of building information models (BIM) in the construction industry has created more opportunities for automated compliance checking (ACC). BIM enables sharing of digital building design data that can be used for compliance checking with legal requirements, which are conventionally conveyed in natural language and not intended for machine processing. Creating a computable representation of legal requirements suitable for ACC is complex, costly, and time-consuming. Large language models (LLMs) such as the generative pre-trained transformers (GPT), GPT-3.5 and GPT-4, powering OpenAI's ChatGPT, can generate logically coherent text and source code responding to user prompts. This capability could be used to automate the conversion of building regulations into a semantic and computable representation. This paper evaluates the performance of LLMs in translating building regulations into LegalRuleML in a few-shot learning setup. By providing GPT-3.5 with only a few example translations, it can learn the basic structure of the format. Using a system prompt, we further specify the LegalRuleML representation and explore the existence of expert domain knowledge in the model. Such domain knowledge might be ingrained in GPT-3.5 through the broad pre-training but needs to be brought forth by careful contextualisation. Finally, we investigate whether strategies such as chain-of-thought reasoning and self-consistency could apply to this use case. As LLMs become more sophisticated, the increased common sense, logical coherence and means to domain adaptation can significantly support ACC, leading to more efficient and effective checking processes.

___________________________________________________________________________

## 1. Introduction

The increasing adoption of building information models (BIM) in construction projects creates more opportunities for automation to boost productivity, minimise duplications of effort, and ultimately improve the quality of buildings and provide a safer and more comfortable living environment. Automation in regulatory compliance checking has been an active research field in the industry for half a century (Dimyadi and Amor, 2013). BIM enables project stakeholders to share digital building data, which is the subject of the automated compliance checking (ACC) process. Legal requirements, the standard to which BIM data must be checked against, are conventionally conveyed in natural language, not intended for machine processing. Having a digital representation of legal requirements is essential for the ACC process. Manually converting regulatory clauses has been considered in many cases (Fenves, 1966; Beach and Rezgui, 2018; Dimyadi et al., 2020), but expertise in multiple domains, such as knowledge engineering, law and construction, is required to achieve high-quality translations. Considering the high number of relevant normative documents for compliance checking (e.g., there are over 600 building-related standards in New Zealand), this process is very time-consuming and costly.

Consequently, natural language processing is a promising direction to automate this task. However, interpreting natural language regulations and converting them into a computer-processable representation poses significant challenges. Complex legalese, ambiguities, and other linguistic intricacies make this task inherently difficult. A common strategy to reduce the complexity is to extract semantic information elements first (Zhang and El-Gohary, 2016; Xu and Cai, 2019) and transform these elements into the final representation in a second step (Zhang and El-Gohary, 2015). Early approaches used hand-crafted rules based on linguistic features and ontologies (Zhou and El-Gohary, 2017). While such rule-based approaches can achieve high accuracy for narrow domains (Zhang and El-Gohary, 2019), they lack scalability and rely on the quality and availability of the ontologies. Machine learning (ML) can be used to circumvent the problems of scalability and manual efforts for crafting rules. Li et al. (2020) and Wang and El-Gohary (2022) extracted entity-relation triplets, and Y. Zhou et al. (2022) conducted semantic labelling using deep neural networks. Nevertheless, human effort is still required to prepare training data for each ML task.

To reduce this effort, Fuchs et al. (2022) used T5 (Raffel et al., 2020), a pre-trained transformer model (Vaswani et al., 2017), to translate building regulations into a formal representation. Transformer is a neural network architecture proposed by Vaswani et al. (2017), which relies heavily on self-attention to generate representations of text. The original transformer architecture is especially suited for translation and semantic parsing tasks since it consists of an encoder to process the input text and a decoder to generate the output while paying attention to specific parts of the input. Through unsupervised pre-training, the model requires less training data and has a better understanding of natural language and common sense. T5 includes further tasks such as translation and summarisation into the pre-training by specifying task prefixes in natural language. By treating the conversion of the regulations into LRML as an end-to-end semantic parsing task, the existing formal representation of New Zealand's building regulations (Dimyadi et al., 2020) could be used for the training rather than having to annotate entities for information extraction. Furthermore, newly translated regulations can be utilised to refine the model and improve the translation quality over time. Nevertheless, this process is strongly dependent on the quality and quantity of the training data (Fuchs et al., 2023a), and it can be difficult to get started with the translation process, especially if there are no earlier efforts.

Large Language Models (LLMs), including generative pre-trained transformers (GPT) (Radford et al., 2018), such as GPT-3.5 by OpenAI, which powers the standard version of ChatGPT[1], Bard[2] by Google, and open-source initiatives like Large Language Model Meta AI (LLaMA) (Touvron et al., 2023) and Alpaca (Taori et al., 2023) have exhibited impressive capabilities in a wide range of tasks without explicit training. These capabilities can be related to the model size and the amount of data used for the pre-training. During inference, tasks are presented to the model using an input prompt, possibly accompanied by a limited set of exemplars. These processes are called in-context and few-shot learning. Using such approaches offers the advantage of initiating the translation of building regulations with as few as five exemplars.

In this study, we aim to evaluate the performance of LLMs, specifically GPT-3.5, in generating a formal representation of building regulations. Results by D. Zhou et al. (2022) and Drozdov et al. (2022) for related semantic parsing tasks indicate that even a small number of exemplars is sufficient to generate the desired syntactical structure. Nevertheless, challenges arise from the extensive vocabulary, complex expressions, and the need to adhere to specific encoding guidelines.

To address these challenges, we have investigated a robust sampling strategy inspired by the work of Drozdov et al. (2022). This strategy helps us identify samples that closely resemble the clause to be translated, resulting in enhanced performance compared to the supervised T5 model. Additionally, we have explored the potential benefits of leveraging the self-consistency technique proposed by Huang et al. (2022) to improve our approach further. Moreover, we have investigated the feasibility of utilising GPT-3.5 as a teacher model to enhance the performance of less resource-intensive models trained under supervision using GPT-generated data. This investigation aims to leverage the knowledge and capabilities of GPT-3.5 to improve the translation performance of these models, despite their limited resources.

Overall, this research contributes to the advancement of automated compliance checking regarding the automated interpretation of building regulations. We harness the power of LLMs to translate natural language legal requirements into formal representations and provide valuable insights into the performance and challenges associated with this approach, paving the way for future improvements and enhanced compliance automation.

## 2. Methodology

### 2.1. LegalRuleML Dataset

LegalRuleML (LRML) is an XML-based (i.e., Extensible Markup Language) formal representation of legal requirements, which was evaluated for representing building regulations by Dimyadi et al. (2017). The representation helps to differentiate between common phenomena in legal text, such as term definitions (i.e., *"ConstitutiveStatement"*) and normative requirements (i.e., *"PrescriptiveStatement"*). Statements are modelled as conditions (i.e., *"if"*) and conclusions (i.e., *"then"*). For prescriptive statements, deontic operators (e.g., *"obligation"*, *"permission"*, *"prohibition"*) are used within the conclusion. The actual domain logic is modelled as conjunctions and disjunctions of boolean statements. Further aspects, such as defeasibility and linkages to the source text and ontologies, are well supported.

This study utilises the corpus of building regulations with their representation in LRML from Fuchs et al. (2022). This dataset was derived from clauses of the New Zealand Building Code translated into LRML by Dimyadi et al. (2020). To make the dataset usable for semantic parsing, the LRML representation was compacted by replacing the XML tags with brackets and removing recoverable information. It was then aligned with the relevant parts of the regulatory clauses. In this study, we utilise the newest version of this dataset by Fuchs et al. (2023b), including the reversible intermediate representation (IR) seen in Listing 1 instead of the original, more verbose representation as the prediction target. Using this IR simplifies the semantic parsing task while a loss-less transformation into the original representation is ensured.

> *Source: 3.4.2 The floor waste shall have a minimum diameter of 40 mm.*
>
> *Original: if(expr(fun(exist),atom(var(floorWaste)))),then(obligation(expr(fun(greaterThanEqual),atom(rel(diameter), var(floorWaste))),data(baseunit(prefix(milli),kind(metre)),value(40.0)))))*
>
> *IR: if(exist(floorWaste)),then(obligation(greaterThanEqual(floorWaste.diameter,40 mm)))))*

**Listing 1.** LRML dataset example

To allow backwards compatibility during the evaluation, the IR is back-translated to the original LRML representation, where an F1-Score for partial entities and their relations is calculated (see Fuchs et al. (2022) for a detailed description). This score provides a holistic quantification of the similarity between predicted and ground truth LRML rules. An emphasis is given to the correctly predicted entities and relations between the entities. All logical operators, such as *"if"*, *"then"*, *"and"*, *"or"*, *"obligation"*, and *"permission"*, are scored based on the correct child entities and relations. Additionally, we evaluate the translation quality of the IR directly by calculating the bilingual evaluation understudy (BLEU) (Papineni et al., 2002), an established translation evaluation metric calculating the n-gram overlap between the prediction and the ground truth.

---

[1] https://chat.openai.com/

[2] https://bard.google.com/

## 2.2. In-context Learning

Large language models (LLMs) are highly receptive to the input prompt. Based on the information provided in the context, the model can learn to perform a wide range of tasks. The model outputs can vary significantly depending on how well one formulates the task, what context is provided, and what and how many exemplars are given. In this light, LLMs are limited by their context length. For example, GPT-3.5 could handle 4,096 tokens at the time the experiments were conducted and GPT-4 (OpenAI, 2023) 8,192 or 32,768 tokens depending on the model version. A token refers to sub-words, and the 4,096 tokens correspond to approximately 3,000 words. The greater length of the LRML rules in our dataset compared to many other semantic parsing datasets limits the number of exemplars to around 30.

We evaluate a range of prompting strategies using GPT-3.5. The following strategies are investigated: 1) Sampling, 2) Contextualisation, 3) Chain-of-thought reasoning, 4) Self-consistency, and 5) GPT as a teacher model.

### 2.2.1. Sampling

First, we test different sampling strategies – Random, hand-picked, diversity, and representative sampling. Samples are provided in a "*Source/Target*" format, as shown in Listing 2. For improved tokenisation, spaces are inserted, and the camel case notation is reverted, as in Fuchs et al. (2023a). The last *"Source"* represents the new clause to translate, and *"Target"* indicates to the model to produce the corresponding LRML rule.

---

*Source: CAS2 2.1.1 The floor area of firecells shall be limited in accordance with Table 2.1.*
*Target: if( exist( firecell)), then( obligation( as per( firecell. floor area, nzbc cas2 t2.1)))*

*Source: G12AS1 6.3.1 Electric and gas storage water heaters shall have their temperature controlled by a ...*
*Target: if( is( storage water heater. type, or( electric, gas))), then( obligation( loop( for each( storage water heater ...*

…

*Source: E2AS1 Lead sheet flashings shall: a) Comply with AS 1804, and b) Have a minimum unit mass of 17 kg/m2.*
*Target:<LLM prediction>*

---

**Listing 2.** In-context learning example

For the random sampling strategy, we select different sets of random samples within the maximum allowable context length. This context length is identified by appending the longest validation clause with its LRML translation to the exemplars and determining the number of tokens required. Comparing different exemplars, which are summing up to the maximum context length, can not only give indications about the variations between different selections but also about the suitability of longer or shorter exemplars.

We manually selected a range of exemplars to test the human judgement of good and representative exemplars. We reviewed the samples of the training set in two iterations. First, we selected subjectively good and representative samples with little overlap. Second, we picked a selection out of those samples that cover a broad range of topics, different clause lengths and LRML constructs. We randomly select different numbers of samples to evaluate different scales of few-shot learning with values of: 1, 3, 5, 10, 20, and 30.

Common strategies for retrieving valuable samples can be found in active learning research (Monarch, 2021). We test two strategies: diversity sampling and representative sampling. For diversity sampling, the training set is clustered based on semantic similarity between either the clauses or the LRML rules using the Sentence Bidirectional Encoder Representations from Transformers (S-BERT) library[3]. The exemplars are then sampled from all the clusters. Alternatively, we use stratified sampling with the different Acceptable Solutions documents as populations. To retrieve representative samples, the most representative sample for each validation clause was calculated using semantic similarity, and the maximum number of those clauses was used as exemplars. Instead of using the same fixed set of exemplars for all predictions, we also retrieve a different set of exemplars for each clause in the validation set. These exemplars are retrieved either with semantic similarity or similar to Drozdov et al. (2022). For the second case, we sort all clauses based on n-gram similarity with the clause in the validation set. We pick the clause with the highest n-gram overlap and continue with the following clause with different n-grams overlapping. When no new n-grams overlap, we start the process from the beginning with a different $n \in \{1, 2, 3\}$. This process ensures that we have exemplars covering all aspects of the validation set, as far as such clauses exist in the training set.

### 2.2.2. Contextualisation

To support the model further with the translation, we provide a description of the translation task, with the relevant syntax, special cases, and the most common relations and entities. While such information could be inferred from the exemplars, we evaluate if a task description leads to superior understanding. In particular, it is of interest whether such a description can reduce the number of samples required for the task. This is examined by comparing the few-shot learning experiment with hand-picked exemplars in Section 2.2.1 with similar runs having additional context.

---

[3] https://www.sbert.net/examples/applications/clustering/README.html

We start this by describing the task in general, including phrases such as *"You are an expert in knowledge engineering, law, and construction."*, which tend to increase the quality of outputs (Salewski et al., 2023). This is followed by a descriptive specification of the format, including LRML keywords and the structure of expressions, highlights of important and common concepts, such as numbers with units, legal references, and a description of how to define variables. Finally, a list of the most common terms in the training set was provided to help the model decide on the granularity of entities and show the range of available predicates.

### 2.2.3. Chain-of-thought prompting

A generally successful method to increase performance and elicit reasoning is chain-of-thought prompting. This method provides a range of exemplars that show step-by-step descriptions of how one might get to the solution (Wei et al., 2022). Similar behaviour was observed by simply adding *"Let's think step by step:"* at the end of the prompt (Zhang et al., 2022). An example of chain-of-thought prompting for LRML is shown in Listing 3.

```
Source: CAS2 2.1.1 The floor area of firecells shall be limited in accordance with Table 2.1.
Let's think step by step:
The firecell is the only element of the condition: exist( firecell)
There is a shall: obligation
The floor area of the firecell is the subject of the obligation: firecell. floor area
The floor area shall be according to the table: as per
The table document is not specified, so we use the clause document: nzbc cas2 t2.1
Target: if( exist( firecell)), then( obligation( as per( firecell. floor area, nzbc cas2 t2.1)))
```

**Listing 3.** Chain-of-thought example

Since the exemplars are lengthy, only a limited number can be provided and especially for long clauses, we need to ensure that the reasoning steps fit into the context length. Accordingly, a shorter decomposition, which consists only of the alignment between natural language phrases and LRML concepts or expressions, is tested, as shown in Listing 4. Additionally, we evaluate a mixture of chain-of-thought and normal exemplars for the superior strategy.

```
Source: CAS2 2.1.1 The floor area of firecells shall be limited in accordance with Table 2.1.
firecells: exist( firecell)
shall: obligation
floor area of firecells: firecell. floor area
in accordance: as per
CAS2 Table 2.1: nzbc cas2 t2.1
Target: if( exist( firecell)), then( obligation( as per( firecell. floor area, nzbc cas2 t2.1)))
```

**Listing 4.** Alignment-based chain-of-thought prompting

### 2.2.4. Self-consistency

Self-consistency (Huang et al., 2022) and self-reflection (Shinn et al., 2023) are common strategies for improving neural networks for natural language processing tasks. Self-consistency dates back to ensemble methods (Dietterich, 2000), where multiple outputs from differently trained models are sampled, and the best prediction is selected per majority vote. In contrast, self-reflection points to the model's ability to evaluate its outputs. For example, Chen et al. (2023) let the LLM describe generated source code line by line and judge the generated source code. Then, they feed this information back into the model for self-debugging purposes.

Since majority votes are more suited for classification tasks with a limited output space, we propose a strategy similar to Drozdov et al. (2022). We generate multiple predictions with different samples and sampling strategies. In the second step, we feed both the source text and the predictions into the LLM and let it pick the best translation. To introduce the task to the model, we provide in-context exemplars where the source text is followed by different predictions and the prediction with the best F1-Score (i.e., Option 2 in Listing 5) is the target. Alternatively, we test the model's capability to interpolate between the predictions using the ground truth as the target of the exemplars. One example of the second strategy can be seen in Listing 5. For the first strategy, the target would be replaced by the best scoring option (i.e., Option 2). An important question for self-consistency is how the predictions should be generated and whether predictions with high scores or greater diversity are more beneficial. We will examine this aspect by using the best three predictions of the previous experiments on the one hand and a more diverse set of predictions with different sampling or prompting strategies on the other hand.

```
Source: E1AS1 The change in direction of a drain shall not exceed 90° at any point.
if( exist( drain)), then( obligation( and( for( drain. direction change), less than equal( drain. direction change, 90°))))
if( has( drain, change in direction)), then( obligation( less than equal( drain. angle of change, 90 deg)))
if( exist( drain)), then( obligation( not( exceed( drain. change in direction, 90 deg))))
Target: if( has( drain, change in direction)), then( obligation( less than equal( change in direction, 90 deg)))
```

**Listing 5.** Self-consistency with interpolation exemplar. Option 1: 38.1%, Option 2: 80.7%, Option 3: 58.2% F1-Scores.

### 2.2.5. GPT as a teacher model

Being able to translate regulations into formal representations with only a few exemplars offers the opportunity to start a semi-automated translation process with little time, cost, and effort. Inspired by Wang et al. (2022), we test the use of LLM for generating training data in two setups. First, we use the assumption that we do not have any training data except the best-performing exemplars from Section 2.2.1. All remaining samples in the training data are newly translated using the LLM with those best-performing exemplars for in-context learning. We then train a T5 model with the best-performing exemplars plus the LLM-translated training data.

In comparison, we use this method for extending our existing training set with up to 855 new training samples with LLM-generated LRML rules. This setup has the benefit of having a preferable ratio between manually and automatically created data. Furthermore, it allows us to use representative sampling per clause, which is expected to yield superior-quality translations.

### 3. Results

Unless explicitly stated otherwise, all experiments are conducted using GPT-3.5-turbo, the model powering the standard version of ChatGPT. We send requests to this model via the OpenAI API[4]. We changed the decoding temperature to 0 to get the most likely output given the prompt, which leads to more factual but less creative outcomes. All prompts were passed in with the user role, which performed better than the system role. Additionally, we ensure that the LRML rule can be extracted from the response (i.e., the response contains *"if("*). If not, we regenerate the translation with increasing temperatures until we receive a suitable response.

### 3.1. Sampling

A natural and straightforward way of picking exemplars is randomly selecting samples from the training set. We randomly sample 50 exemplars and reduce the number until they fit into the maximum token length. By repeating this process multiple times, we can gain insight into the variance of different sets of exemplars and the number of exemplars while keeping the number of input tokens approximately the same. While the results in Fig. 1 show that the F1-Scores range between 53.5% and 60.6%, the majority of runs were above 57.1%. A weak positive correlation (i.e., 0.275 Pearson's Correlation) between F1-Scores and the number of exemplars can be noticed, but this result is not significant (i.e., 0.441 P-Value), and more data would be required to draw any conclusions.

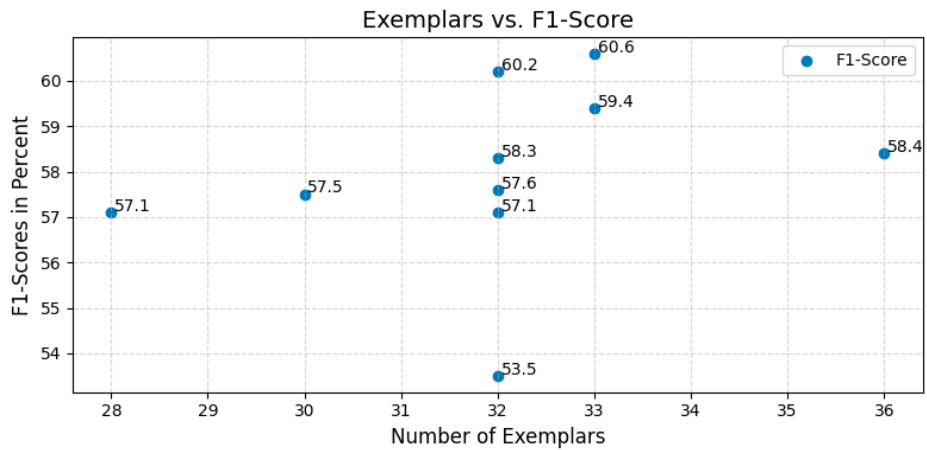

**Fig. 1.** Random sampling and the importance of meaningful exemplars

We manually filtered the training samples to include a diverse and high-quality set of exemplars to evaluate if human judgement benefits the exemplar selection. Since the remaining 45 exemplars represent the training data well, we use this set to investigate how many exemplars are necessary for achieving satisfactory results. We randomly selected 30 exemplars fitting into the context window and reduced the exemplars gradually. Fig. 2 indicates that three exemplars are enough to make meaningful predictions. Still, the model did not stop consistently after predicting the LRML rules causing the low BLEU scores, which got resolved after having five or more exemplars. Having better scores with ten exemplars than with 15 exemplars shows that adding more exemplars does not necessarily improve the performance, especially if such exemplars are not representative of the ground truth data. Nevertheless, a consistent upward trend in F1-Scores can be observed, reaching a peak of 57.8%. This value falls within the range of random sampling results and suggests that human selection does not yield better results for this particular use case.

We investigate strategies to either diversify the exemplars or increase their similarity to the validation data. The results are presented in Table 1. Diversifying the exemplars via stratified sampling per document had adverse effects, likely because, for many documents, there are only a few samples, and those are not representative of other documents. In contrast, clustering the training data based on clauses or LRML rules yielded results that fell within the higher range of the random sampling outcomes depicted in Fig. 1. While representative sampling for the entire validation set was counterproductive, retrieving representative exemplars for each clause in the validation set individually brings significant improvements. The proposed n-gram-based retrieval of exemplars outperforms the semantic-based method and even the T5 model with supervised learning, which had an F1-Score of 68.4% on the validation set using the evaluation procedure from Fuchs et al. (2023b). Furthermore, it can be noticed that having the most similar exemplars at the end of the prompt, i.e., closest to the prediction, brings the best results.

---
[4] https://platform.openai.com/

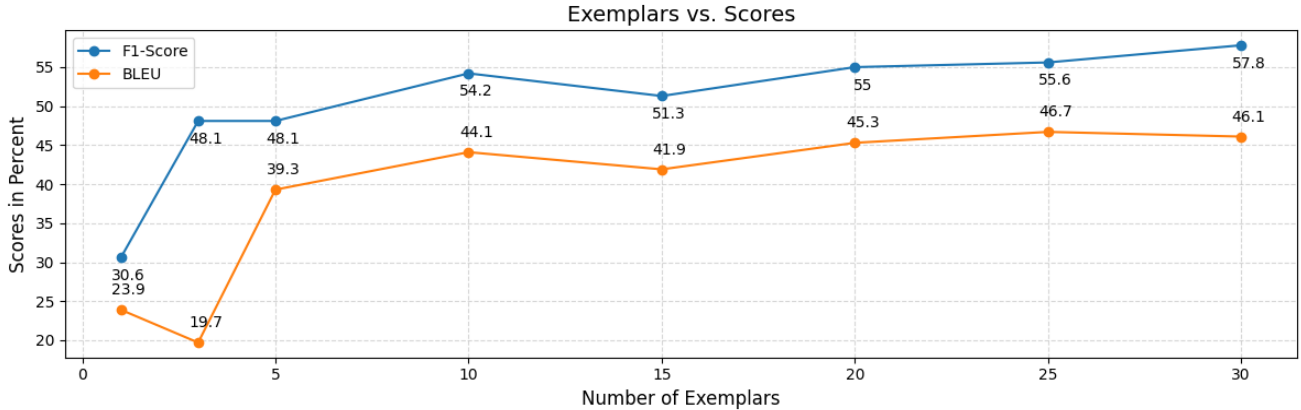

**Fig. 2.** Gradual increase of manually filtered exemplars

**Table 1.** Sampling strategies. Number of Exemplars were averaged for per clause sampling.

| Sampling Strategy | Number of Exemplars | BLEU | F1-Score |
| --- | --- | --- | --- |
| Stratified Sampling | 31 | 46.0% | 56.2% |
| Semantic Clustering Clause | 27 | 45.8% | 59.2% |
| Semantic Clustering LRML | 32 | 50.2% | 59.7% |
| Representative Sampling (RS) | 27 | 47.6% | 55.8% |
| RS per Clause (semantic) | 30.8 | 56.4% | 66.4% |
| RS per Clause reversed (semantic) | 30.8 | 55.6% | 68.3% |
| RS per Clause (n-grams) | 30.2 | 58.2% | 69.1% |
| RS per Clause reversed (n-grams) | 30.2 | 59.3% | 70.3% |

### 3.2. Contextualisation

Due to the limited context length, there is a trade-off between the additional context information and the resulting decrease in the number of exemplars. We evaluated this trade-off with the manually filtered samples in Section 3.1. Including a task introduction reduced the exemplar number from 30 to 28. The number of exemplars was further reduced to 27 by adding the format specification, to 26 by adding the explanation for references, and to 24 by listing the most common terms. This setup can be contrasted with having only ten in-context exemplars, and contextualisation is additional information. Lastly, we combine contextualisation with representative sampling per clause in reverse order. Since the most representative exemplars will be selected first, reducing the number of exemplars might have less effect. As shown in Fig. 3, adding a short task introduction improved all scores continuously over all setups. The specification and reference examples were counterproductive in all setups. This could be rooted in a poor wording, which confuses the model more than it helps. Providing the most common terms and emphasising the important aspects led to higher F1-Scores for predictions with fewer exemplars. Notably, the BLEU scores increased for all setups, which might be related to the low baseline compared to Table 1. Finally, contextualisation with representative sampling per clause had similar improvements of 0.8% F1-Scores with only the introduction and with the full context but higher BLEU scores with the full context. Despite overall mixed results, we believe LLM should be contextualised succinctly to maximise their potential.

### 3.3. Chain-of-thought Prompting

While chain-of-thought (CoT) prompting was a successful strategy to elicit reasoning and improve accuracy in recent works, including semantic parsing, Table 2 suggests this is not the case for generating the long and complex LRML rules. This was confirmed in multiple setups, where the ten manually filtered exemplars in Section 3.1 were reformulated as CoT exemplars in a colloquial language and an alignment-based style. Additionally, we experimented with providing the additional exemplars without CoT before the CoT exemplars to provide overall more exemplars but still enforce CoT reasoning. Finally, we provided additional context together with the CoT exemplars. While these strategies brought slight improvements, the results were consistently worse than when predicting LRML directly. Different methods to formulate the CoTs and recursive prompting strategies should be tested to confirm these results.

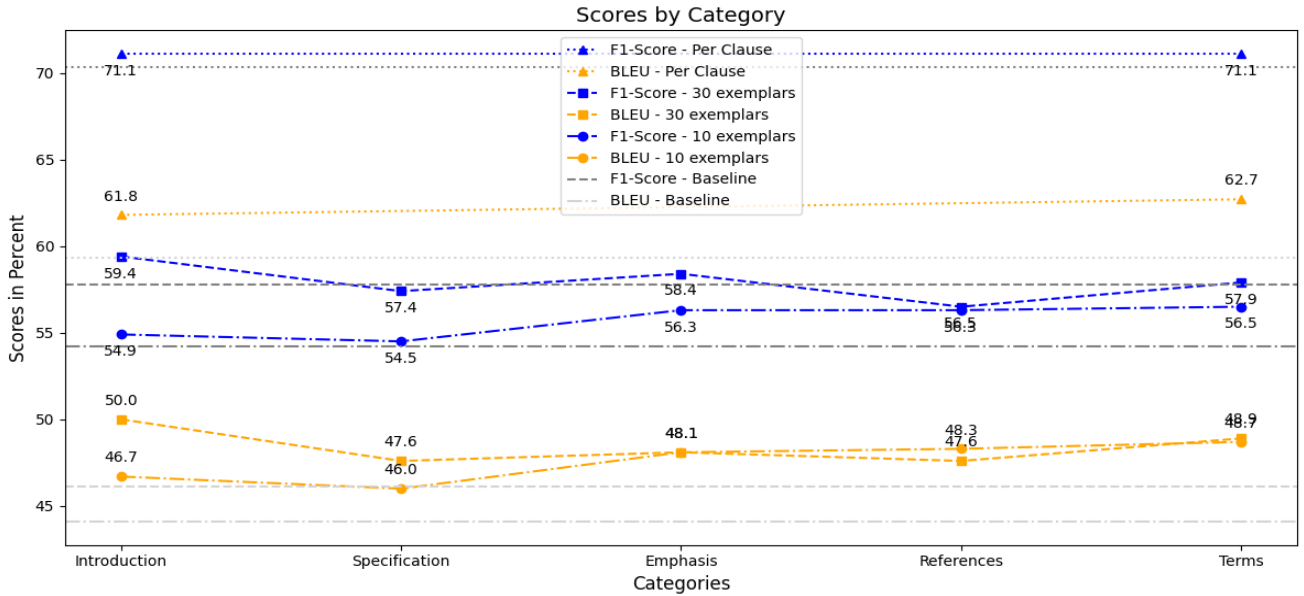

**Fig. 3.** Contextualisation

**Table 2.** CoT prompting

| Experiment | Number of CoT Exemplars | Number of Additional Exemplars | BLEU | F1-Score |
|---|---|---|---|---|
| CoT | 10 | - | 36.8% | 49.0% |
| Additional Exemplars + CoT | 10 | 8 | 42.1% | 52.4% |
| Additional Exemplars + CoT (5) | 5 | 19 | 43.2% | 55.3% |
| CoT Alignment-Based | 10 | - | 39.1% | 50.0% |
| CoT Alignment-Based + Context | 10 | - | 42.5% | 53.0% |

### 3.4. Self-consistency

In the proposed self-consistency strategy, the model is required to reflect on previous outputs and choose the best one. While self-reflection was argued to be an emergent ability related to the size of LLM, in our experimental setup, it could also be ascribed to the intuition that discrimination between different predictions is a simpler task than generating a new LRML rule. The results presented in Table 3 indicate such a behaviour only in one of the setups. Using the three best runs with representative sampling per clause with and without contextualisation (71.1%, 71.1%, and 70.3% F1-Scores) and teaching the model to pick one of the LRML rules yielded an improvement of 0.9% F1-Score outperforming previous supervised approaches using intermediate representations and self-reflection (Fuchs et al., 2023b). Nevertheless, comparing these results to the Oracle F1-Score of 75.4% indicates that the model makes more often than not non-optimal choices. All other setups have negative consequences indicating that high variability in the quality of the LRML options is not helpful. We picked the best runs with representative sampling per clause (71.1% F1-Score), random sampling (60.6% F1-Score), and clustering (59.7% F1-Score) for the mixed setup, but the results were closer to the average instead of bringing improvements. Also, letting the model interpolate between the LRML options yielded worse results than choosing one option.

**Table 3.** Self-consistency

| Experiment | Max F1-Score | Average F1-Score | Oracle F1-Score | BLEU | F1-Score |
|---|---|---|---|---|---|
| Self-consistency – Predict from Mixed | 71.1% | 63.8% | 72.7% | 50.2% | 60.4% |
| Self-consistency – Choose from Mixed | 71.1% | 63.8% | 72.7% | 53.9% | 63.4% |
| Self-consistency – Predict from Best | 71.1% | 70.8% | 75.4% | 54.6% | 64.2% |
| Self-consistency – Choose from Best | 71.1% | 70.8% | 75.4% | 62.3% | 72.0% |

### 3.5. GPT as a Teacher Model

By using GPT as a teacher model, we determine if LLMs can be used to commence training a semantic parser for building regulations with only little effort. In the first setup, we selected the best run from the previous experiments with a fixed set of exemplars, i.e., random sampling with 60.6% F1-Score. By using these exemplars, we generate LRML rules for all remaining clauses in the training set (i.e., 543 LRML rules). We trained the more lightweight and freely available T5 model for semantic parsing using the newly generated training data. We followed the evaluation procedure described in Fuchs et al. (2023b). We report the results for the random validation set and test set in Table 4, which makes results comparable with the earlier research as well as with this work. BLEU scores were dropped since they were calculated based on the LRML representation reversed to the original (see Section 2.1) rather than the intermediate LRML representation, as done in this work. It can be noticed that this method achieved reasonable results with a 58.3% F1-Score for the test set. Nevertheless, this score is 8.1% lower compared to training with the original training data. In this case, the quality of the generated training data with a 59.5% F1-Score might be the limiting factor.

In the second setup, we used additional clauses translated with the representative sampling per clause strategy and contextualisation, intending to improve the T5 model's parsing quality. The first 150 clauses were extracted from unseen New Zealand's (NZ) Acceptable Solution documents. Since there was only a negative effect visible, we added 705 more clauses from NZ Standards, which brought an improvement of 1.2% in F1-Scores showing that generated additional training data could help the model performance. However, improvements to the quality of the generated data might make this strategy more helpful. Returning to the hypothesis that discrimination is easier than generation, asking humans to rank different translations for the same clause might be a valuable strategy to improve the generated data's quality.

Table 4. GPT as teacher model

| Experiment | Training Samples | Original F1-Score (Test) | F1-Score (Valid) | F1-Score (Test) |
|---|---|---|---|---|
| Recreate Training Data | 576 | 66.4% | 59.2% | 58.3% |
| Generate Additional Data – 150 | 726 | 66.4% | 67.7% | 66.1% |
| Generate Additional Data – 855 | 1431 | 66.4% | 67.8% | 67.6% |

## 4. Discussion

The results in this paper show the suitability of LLM to support the translation of building regulations into formal representations insofar as higher scores could be reached compared to traditional supervised models. Additionally, such supervised models could be augmented with additional training data. Nevertheless, since the BLEU and F1-Scores only act as an approximation to how close the translation is to the given ground truth translation, we examine some of the predictions and propose directions for future work.

We recognise that slight changes to the rules can cause significant penalties in F1-Scores. Furthermore, some variations in the predictions, such as different granularities, different terminologies, or alternative expressions, might not impact the actual use cases of the formal representation. For example, the prediction in Listing 6 uses *"has"* instead of *"include"*, and misses the *"and"* keyword to indicate the conjunction. While the *"and"* would reduce the F1-Scores by about 1.2%, the *"has"* has a stronger impact of 9.7% since it also influences the scoring of the logical and deontic keywords, *"then"* and *"permission"*.

---
*Source:*     G13AS2 5.9.3 Access points may be located in a space containing a soil fixture.

*Prediction:* if( exist( access point)), then( permission( within( access point. location, space), has( space, soil fixture)))

*Target:*     if( exist( access point)), then( permission( and( within( access point. location, space), include( space, soil fixture))))

---
**Listing 6.** Minor mistakes. 89.1% F1-Score

Listing 7 shows an example where *"ventilation pipe"*, a term from the LRML vocabulary, was used in the ground truth rather than *"vent pipe"*. Furthermore, *"vent pipe"* was accessed directly through *"storage water heater"* in the prediction, while the target rule established this relation explicitly. While semantically correct, the predicted LRML's F1-Score was decreased by over 20% F1-Score. Some of these issues could be avoided by having all necessary terms in the exemplars or training data, having the LRML vocabulary accessible, or switching to a semantic or more tolerant evaluation metric.

---
*Source:*     G12AS1 6.3.2 Open vented storage water heaters shall have a vent pipe complying with Paragraph 6.8.

*Prediction:* if( is( storage water heater. type, open vented)), then( obligation( comply with( storage water heater. vent pipe, nzbc g12as1 6.8)))

*Target:*     if( is( storage water heater. type, open vented)), then( obligation( and( has( storage water heater, ventilation pipe), comply with( ventilation pipe, nzbc g12as1 6.8))))

---
**Listing 7.** Vocabulary alignment and implicit relations. 79.1% F1-Score

While most inconsistencies in the LRML dataset were addressed by Fuchs et al. (2023a), it can be problematic to identify when information in parenthesis (i.e., *"valve vented"*) is relevant and should be encoded (see Listing 8). In comparison to previous work (Zhou and El-Gohary, 2017), where such information was removed entirely, this is part of the complexity of the LRML dataset. In addition, to calculate the F1-Score, the intermediate LRML representation is reversed to its original form, which includes restoring triplets and quadruplets which were merged: e.g., *"or( is( storage water heater. type, unvented), is( storage water heater. type, valve vented))"*. Accordingly, the missing *"valve vented"* leads to a much greater reduction in F1-Scores.

> *Source:*     *G12AS1 6.10.1 NZS 4607 is an acceptable solution for unvented (valve vented) storage water heaters, but may exceed the performance criteria of NZBC G12.*
>
> *Prediction:* *if( is( water heater. type, unvented)), then( permission( comply with( water heater, nzs 4607)))*
>
> *Target:*     *if( is( storage water heater. type, or( unvented, valve vented))), then( permission( comply with( storage water heater, nzs 4607)))*

**Listing 8.** F1-Score calculation impact. 70.6% F1-Score

Another common difficulty is deciding where to draw the entity boundaries and when to split an entity into a subexpression. While in the example in Listing 9, expressing *"opening window"* as a *"window"* with property *"openable"* is the better choice, this could still be a valid encoding depending on the information alignment in the compliance checking use case. Accordingly, the best way to properly evaluate this would be by calculating the execution accuracy in an end-to-end compliance checking system.

> *Source:*     *G15AS1 Waste Storage Area 3.0.6 opening windows shall be screened to prevent entry by insects and other vermin.*
>
> *Prediction:* *if( has( waste storage area, opening window)), then( obligation( screened from( opening window, and( insect, vermin))))*
>
> *Target:*     *if( and( has( waste storage area, window), is( window, openable))), then( obligation( screened from( window, vermin)))*

**Listing 9.** Different granularity. 67.1% F1-Score

Depending on the selection of the relation, the subject and object can be swapped, which causes additional F1-Score penalties. A reduction of 26.1% F1-Score can be noticed in Listing 10 despite the high semantic similarity.

> *Source:*     *B1AS1 Structure Design 8.0 Small Chimneys; See Acceptable Solution B1/AS3.*
>
> *Prediction:* *if( and( is( structure, chimney), is( chimney. size, small))), then( apply to( nzbc b1as3, structure))*
>
> *Target:*     *if( and( is( structure, chimney), is( chimney. size, small))), then( as per( chimney. design, nzbc b1as3))*

**Listing 10.** Semantically similar translations. 73.9% F1-Score

Although there are some logical errors in the prediction in Listing 11, this example shows the model's difficulty in learning the syntax required to compare particular objects. Defining variables within the logical consequence and using them in string values is an unusual way of encoding the information, which is challenging to be trained into the model. Previous experiments conducted with the example in Listing 11 revealed that a more natural approach for encoding calculations involves utilising functions for computations and permitting the utilisation of functions and entities with properties as the second argument of a predicate. An alternative ground truth, which includes those changes, would be *greater than equal( top of( gully trap. water seal), subtract(top of( gully trap. gully dish), 600 mm)).*

> *Source:*     *G13AS2 Drainage 3.3.1 All gully traps shall have (see Figures 2 and 3): The top of the water seal no more than 600 mm below the top of the gully dish.*
>
> *Prediction:* *if( exist( gully trap)), then( obligation( and( has( gully trap, water seal), has( gully trap, gully dish), less than equal( water seal. elevation, 'gully dish. elevation - 600 mm'))))*
>
> *Target:*     *if( exist( gully trap)), then( obligation( and( define( top of( gully trap. gully dish), x0), greater than equal( top of( gully trap. water seal), 'x0 - 600 mm'))))*

**Listing 11.** Syntax difficulties and logical errors. 57.8% F1-Score

Finally, good alternative translations as shown in Listing 12 can lead to the same outcome while scoring extraordinarily low. In such scenarios, traditional semantic similarity metrics are unlikely to be effective, but alternative measures such as execution accuracy and logical equivalence may prove more suitable.

> *Source:*     *G13AS1 Individual floor waste pipes connected to a floor waste stack need not be vented (see Figure 3).*
>
> *Prediction:* *if( and( has( floor waste, pipes), is( connect( pipes), floor waste stack))), then( permission( not( has( pipes, ventilation))))*
>
> *Target:*     *if( part of( pipe, floor waste stack)), then( permission( not( has( pipe, ventilation))))*

**Listing 12.** Reasonable alternative translation. 37.1% F1-Score

By evaluating the in-context learning capabilities of LLMs and investigating a range of example predictions, we show that LLMs can produce high-quality semantic translations of building regulations. The application of LLMs is not restricted to a specific formal representation, but their remarkable sample efficiency can facilitate the translation process into various formal representations. Leveraging the LLM's exposure to widely adopted representations, such as first-order logic, could yield even better parsing results and allow established reasoners to assess execution accuracy for an evaluation closer to the real-world application.

While in-context learning can be an excellent method to utilise LLMs without the need for fine-tuning, semantic parsing into representations with large vocabulary sizes can be problematic. Either the model is forced to make up new relations and entities or must be introduced to all relevant terms. For this task, the model needs to do both, reusing the existing vocabulary as much as possible while being able to define new vocabulary in a manner similar to human translators. This could be achieved by fine-tuning GPT or models such as the LLaMA model by Touvron et al. (2023) with 7 Billion parameters, which has an order of magnitude more parameters than the T5 model used in previous work and can be fine-tuned on a single GPU with 48GB memory with parameter efficient fine-tuning strategies, such as low-rank adaptation (LoRA) (Hu et al., 2021). Alternatively, using retrieval-augmented LLMs to look up the vocabulary or pre-training on the LRML vocabulary and other classification systems could help to inject that knowledge into models such as T5.

## 5. Conclusions and Future Work

This work proposed the use of large language models (LLMs) to automate the translation of building regulations into a formal representation usable for automated compliance checking (ACC). The pre-training of LLMs familiarises them with a large variety of tasks and topics so that only a few exemplars are enough to generate translations with reasonable F1-Scores. The sample efficiency opens up new means to provide support to translators and shows potential for full automation even if no initial training data is available. Furthermore, the sample efficiency gives more freedom in selecting the representation format without dependence on data availability. LLMs can outperform supervised methods by using sophisticated methods to select the most appropriate exemplars for each clause to translate. In addition, small models with supervised learning could be improved further with the help of synthetic training data and limited human supervision, which might be the more feasible solution to apply these techniques in practice.

In recent years, the exploration of prompting techniques for LLMs has gained considerable momentum. Accordingly, a plethora of prompting strategies evolved, which could not all be exhaustively examined in the scope of this study. Notably, advanced techniques like recursive prompting and strategies that involve querying knowledge bases have the potential to be particularly beneficial within the context of the evaluated use case. Furthermore, the examination of various LRML rules reveals that numerous translations exhibit not only logical coherence but also surpass the quality indicated by the F1-Scores. Still, a disparity persists in terms of translation quality suitable for ACC application. Even when nearing flawless translation, the potential requirement for certification and review remains, emphasising the need for a human-assisted workflow.

Building on these conclusions and limitations, our future work takes two paths. Firstly, we will explore the feasibility of involving regulators in translating regulations into formal representations during regulatory drafting and publishing. We will also investigate the types of tool support that could enhance the feasibility of this process. Secondly, we aim to advance the implementation of the resulting rules in real-world situations. This not only enhances the practical usability of these rules but also contributes to refining the translation process. By placing more emphasis on logical correctness in evaluations, we aim to achieve a more dependable assessment and thus enhance the overall quality of the formal representation of building regulations.

**Author Contributions**

Stefan Fuchs contributed to conceptualisation, methodology, implementation, validation, analysis, draft preparation, visualisation, and manuscript editing. Michael Witbrock contributed to conceptualisation, computational resources, manuscript editing, and supervision. Johannes Dimyadi contributed to manuscript editing and supervision. Robert Amor contributed to conceptualisation, methodology, manuscript editing, and supervision. All authors have read and agreed with the manuscript before its submission and publication.

**Funding**

This research was funded by the University of Canterbury's Quake Centre's Building Innovation Partnership (BIP) programme, which is jointly funded by industry and the Ministry of Business, Innovation and Employment (MBIE).

**Ethics Statement**

Applying machine learning to solve legal tasks can have broad implications since the correctness of the algorithm cannot be guaranteed. In the building domain, this could cause major adverse events that put building occupants at risk if an unsafe building is issued a compliance certificate and opened to the public. Nevertheless, this work treats LLMs as a supporting tool, which helps to automate compliance checking rather than executing the compliance checks directly. The entire means of having a formal representation of building codes, which is verifiable and can be deterministically executed, is to guarantee interpretable and reliable automated compliance checking.